\documentclass{article}

\usepackage{microtype}
\usepackage{graphicx}
\usepackage{subcaption}
\usepackage{booktabs} % for professional tables
\usepackage{bbm}

\usepackage[pagebackref]{hyperref}

\usepackage[preprint]{icml2026}

\usepackage{amsmath}
\usepackage{amssymb}
\usepackage{mathtools}
\usepackage{amsthm}
\usepackage{bm}
\usepackage{multirow}
\usepackage{graphicx}
\usepackage{makecell}
\usepackage{color}
\usepackage{wrapfig}
\usepackage{marvosym}
\usepackage{pifont}
\newcommand{\cmark}{\ding{51}}%
\newcommand{\xmark}{\ding{55}}
\usepackage{colortbl}
\usepackage{algorithm}
\usepackage{amsmath,amssymb}

\usepackage[capitalize,noabbrev]{cleveref}

\theoremstyle{plain}
\newtheorem{theorem}{Theorem}[section]
\newtheorem{proposition}[theorem]{Proposition}

\theoremstyle{definition}

\theoremstyle{remark}

\usepackage[textsize=tiny]{todonotes}

\newcommand{\ie}{{\emph{i.e.}}}
\newcommand{\eg}{{\emph{e.g.}}}

\newcommand{\fourier}[1]{\hat{#1}} % <--- 定义在这里

\newcommand{\E}{\mathbb{E}}

\icmltitlerunning{DCCT for Generalizable AI-Generated Image Detection}

\begin{document}

\twocolumn[
  \icmltitle{Color Matters:  Demosaicing-Guided Color Correlation Training for Generalizable AI-Generated Image Detection}



  \icmlsetsymbol{equal}{*}

  \begin{icmlauthorlist}
    \icmlauthor{Nan Zhong}{Nan Zhong}
    \icmlauthor{Yiran Xu}{Yiran Xu}
    \icmlauthor{Mian Zou}{Mian Zou}
  \end{icmlauthorlist}

  \icmlaffiliation{Nan Zhong}{Department of Computer Science, City University of Hong Kong, Hong Kong}
  \icmlaffiliation{Yiran Xu}{Department of Computer Science, Fudan University, Shanghai, China}
  \icmlaffiliation{Mian Zou}{School of Computing and Artificial Intelligence, Jiangxi University of Finance and Economics, Nanchang, China }

  \icmlcorrespondingauthor{Mian Zou}{zoumian@jxufe.edu.cn}


  \vskip 0.3in
]



\printAffiliationsAndNotice{}  % no special notice (required even if empty)

\begin{abstract}
As realistic AI-generated images threaten digital authenticity, we address the generalization failure of generative artifact-based detectors by exploiting the intrinsic properties of the camera imaging pipeline.
Concretely, we investigate color correlations induced by the color filter array (CFA) and demosaicing, and propose a \textbf{D}emosaicing-guided \textbf{C}olor \textbf{C}orrelation \textbf{T}raining (DCCT) framework for AI-generated image detection. By simulating the CFA sampling pattern, we decompose each color image into a single-channel input (as the condition) and the remaining two channels as the ground-truth targets (for prediction). A self-supervised U-Net is trained to model the conditional distribution of the missing channels from the given one, parameterized via a mixture of logistic functions. Our theoretical analysis reveals that DCCT targets a provable distributional difference in color-correlation features between photographic and AI-generated images. By leveraging these distinct features to construct a binary classifier, DCCT achieves state-of-the-art generalization and robustness, significantly outperforming prior methods across over 20 unseen generators.
\end{abstract}

\section{Introduction}
The rapid proliferation and growing sophistication of image generation models~\cite{kingma2016improved,kingma2018glow,goodfellow2014generative,karras2017progressive,dhariwal2021diffusion,gu2022vector,ho2020denoising} democratize not only creative expression but also the capacity for large-scale deception, as they now produce images that are often indistinguishable from photographs~\cite{nightingale2022ai, news}, underscoring the need to reliably distinguish camera-captured photographs from AI outputs~\footnote{Throughout this paper, we refrain using terms ``real'' and ``fake'' when contrasting photographic and AI-generated imagery, as contemporary generative models can memorize and reproduce photographs from their training data~\cite{carlini2023extracting, kadkhodaiegeneralization, zou2024semantic}}.

Over the past few years, a vast body of research on AI-generated image (AIGI) detection~\cite{wang2020cnn,cazenavette2024fakeinversion,luo2024lare,wang2023dire,ma2023exposing,yan2025sanity} has converged on an artifact-based strategy. The core principle is to isolate and analyze generative artifacts, \ie, patterns inherent to the synthesis process that are not found in natural photographic images~\cite{liu2020global, durall2020watch, wang2023dire, corvi2023intriguing,  tan2024rethinking,chendrct, yan2025sanity}. 
However, these methods primarily focus on artifacts specific to a single family of generative models (\eg, reconstruction errors via diffusion inversion~\cite{wang2023dire}); however, as generators evolve, such artifacts tend to shift, leading to reduced detection reliability.
A distinct line of work pursues model-agnostic fingerprints, which has evolved from targeting manually-defined inconsistencies like physiological or physical flaws~\cite{yang2019exposing, guo2022icassp_eyes, hu2021exposing, bohavcek2023geometric} to leveraging general image representations by vision-language pretraining~\cite{ojha2023towards, cozzolino2023raising}, yielding superior performance.
Yet, because their pretraining tasks are not explicitly tailored to AIGI detection, these approaches can still fall short of optimal.

In this paper, we propose a novel \textbf{D}emosaicing-guided \textbf{C}olor \textbf{C}orrelation \textbf{T}raining (DCCT) framework for AIGI detection. DCCT explicitly drives the model to learn the camera-intrinsic color correlations induced by the color filter array (CFA) and demosaicing process~\cite{lukac2005color,ramanath2005color}, which we demonstrate to be a stable, physically grounded cue for separating photographic from AI-generated images.
By simulating the Bayer CFA sampling pattern, we decompose each RGB image into a one-channel observation and the remaining two-channel reconstruction targets, mirroring the RAW-to-RGB interpolation step in photographic demosaicing. Motivated by the CFA aliasing behavior, we operate in the high-pass domain to emphasize subtle inter-channel dependencies. On high-pass filtered residuals~\cite{fridrich2012rich}, a U-Net~\cite{ronneberger2015u} is trained in a self-supervised manner to estimate the conditional distribution of the missing channels given the observed one, parameterized by a mixture of logistic functions in the spirit of PixelCNN++~\cite{salimans2017pixelcnn}. 
From a theoretical standpoint, DCCT guarantees a provable, non-vanishing distributional gap between photographic and AI-generated images in the learned color-correlation feature space, formalized as a uniform lower bound on the 1-Wasserstein distance over CFA aliasing–related high-frequency subbands.

Building on this camera-aware pretraining, we then use the learned color-correlation features as input to a lightweight binary classifier for AIGI detection. Extensive experiments on multiple benchmarks, covering more than 20 unseen generators, show that DCCT delivers state-of-the-art generalization and robustness, consistently outperforming artifact-based and generic representation-based baselines.

In summary, the contributions of this paper include \textbf{1)} A DCCT framework that supports generalizable AIGI detection; \textbf{2)} A theoretical analysis of DCCT,  elucidating its plausible working mechanism that separates photographic and AI-generated images; and \textbf{3)} An extensive experimental demonstration on the superior generalizability and robustness of the proposed DCCT method.

\section{Related Work}

\noindent\textbf{Generative Models.} 
Deep generative models~\cite{kingma2016improved, kingma2018glow, goodfellow2014generative, song2020score, salimans2017pixelcnn} have rapidly evolved, where GANs~\cite{goodfellow2014generative,karras2017progressive,karras2019style,brock2018large} and diffusion models~\cite{ho2020denoising,song2020score} currently dominate high-fidelity image synthesis. 
GANs train a generator to create images from random noise, while a discriminator learns to distinguish them from photographic ones in an adversarial game. Improvements in loss functions and regularization~\cite{arjovsky2017wasserstein, miyato2018spectral, gulrajani2017improved}, together with progressively grown and style-based architectures~\cite{karras2017progressive, karras2020analyzing} and conditional designs driven by labels, text or images~\cite{odena2017conditional, isola2017image}, have enabled sharp, high-resolution, and controllable outputs. 
Diffusion models instead define a forward process that gradually corrupts an image with Gaussian noise and learn a neural network to reverse this process step by step, starting from noise. 
With carefully designed noise schedules and samplers, they now achieve state-of-the-art image quality and diversity~\cite{dhariwal2021diffusion,podell2023sdxl,saharia2022photorealistic}, and have been heavily optimized for faster sampling and richer conditioning~\cite{zheng2023fast, song2023consistency, rombach2022high, ruiz2023dreambooth}.
Despite architectural differences, modern generative models operate purely in the digital/spatial domain, directly synthesizing RGB values via learned dependencies without camera imaging pipeline, whereas camera-captured photographs result from analog light transport, CFA mosaicked sensing, and subsequent ISP operations~\cite{lukac2005color,ramanath2005color}.
This gap leads to systematically different color correlations, which our method exploits.

\noindent\textbf{AIGI Detection.}
The proliferation of AIGIs has catalyzed research into their detection, with a primary approach being the exploitation of artifacts from specific generators. Early methods sought to identify generator-specific artifacts. For example, research on GANs targeted frequency distortions from upsampling~\cite{durall2020watch, corvi2023intriguing, dong2022think, frank2020leveraging}, while recent work on diffusion models has exploited inversion and reconstruction errors~\cite{wang2023dire, ma2023exposing, cazenavette2024fakeinversion, luo2024lare}. While effective in-domain, these methods overfit to the artifacts of known generators and fail to generalize to novel architectures.
A second, more generalizable paradigm involves adapting large-scale pretrained models. Typically, generic features are extracted using a foundation model like CLIP~\cite{ojha2023towards, cozzolino2023raising}, often followed by a fine-tuning stage~\cite{yan2025orthogonal, zou2025bi} to optimize them for distinguishing between photographs and AI-generated images.
However, these features lack an intrinsic understanding of the image formation process, thus limiting their ultimate effectiveness. 
Recognizing this gap, the frontier of research is moving toward camera-aware pretraining, for instance, by using camera metadata~\cite{zou2025self, zou2025bi, zhong2025self} or pixel-correlation tasks~\cite{cozzolino2024zero} to imbue models with knowledge of the photographic process. 
In this paper, we advance this direction from an ISP perspective by explicitly modeling CFA sampling and demosaicing during pretraining, encouraging the feature discrepancies between physical imaging and purely digital generation.

\noindent\textbf{CFA-Based Image Forensics.}
Long before AIGIs, image forensics had already leveraged CFA sampling and demosaicing as physical cues, showing that demosaicing leaves interpolation artifacts that support image forgery detection and localization~\cite{popescu2005exposing,ferrara2012image}. 
More recent studies move from explicit interpolation models to richer CFA-aware feature spaces and ensemble classifiers, achieving robust camera model identification across many devices~\cite{bayram2005source, chen2015camera}. 
In parallel, sensor pattern noise was used as a device fingerprint, with methods leveraging the CFA structure to remove interpolation artifacts for more robust source identification and integrity verification~\cite{li2011color}. 
Building on research that uses stable, camera-specific CFA characteristics to differentiate genuine from tampered content, our work adapts this forensic approach to distinguish camera-captured photographs from AI-generated images.

\begin{figure}
    \centering
    \subfloat[Original scene]{\includegraphics[width=0.33\linewidth]{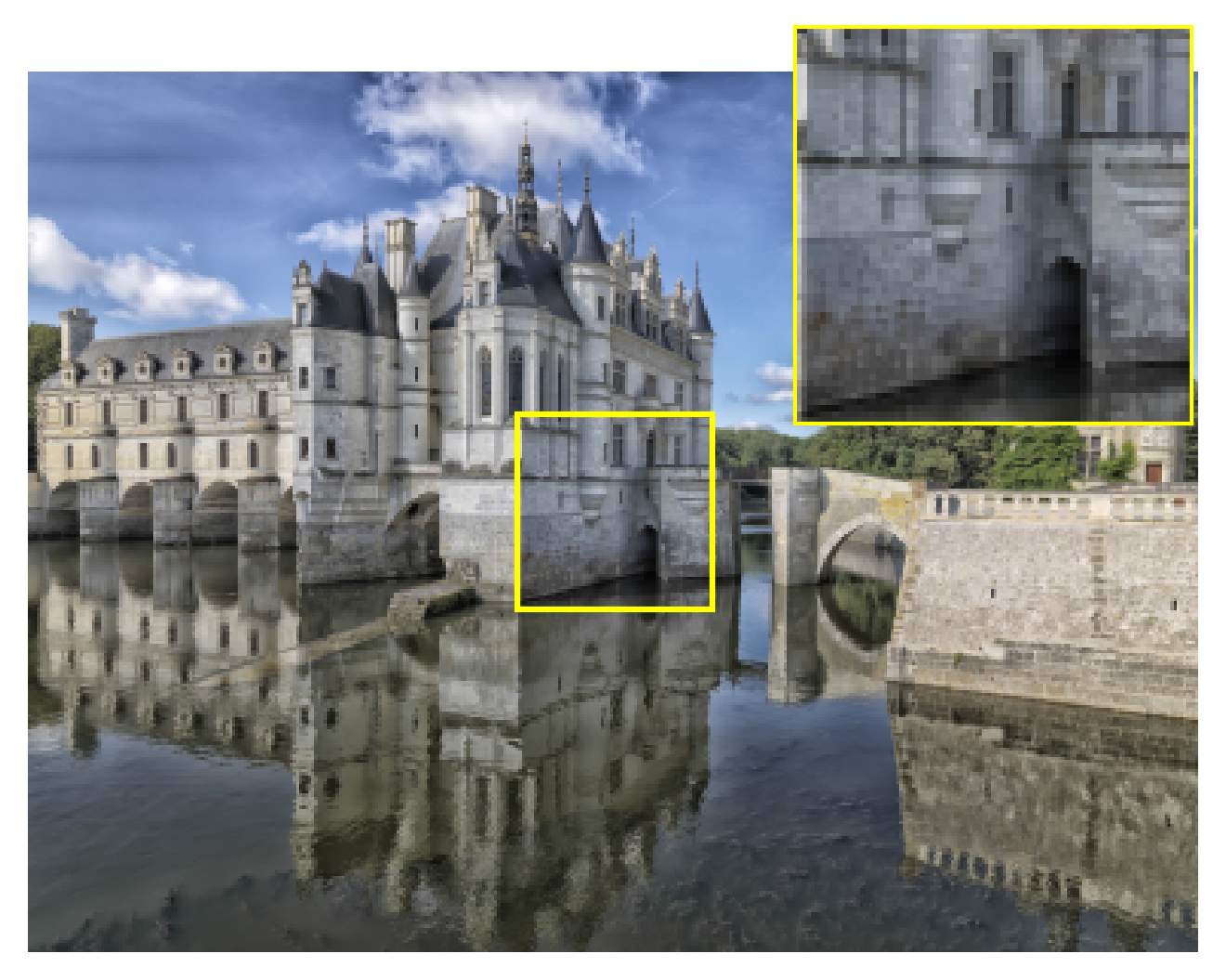}}\hfill
    \subfloat[RAW CFA mosaic]{\includegraphics[width=0.33\linewidth]{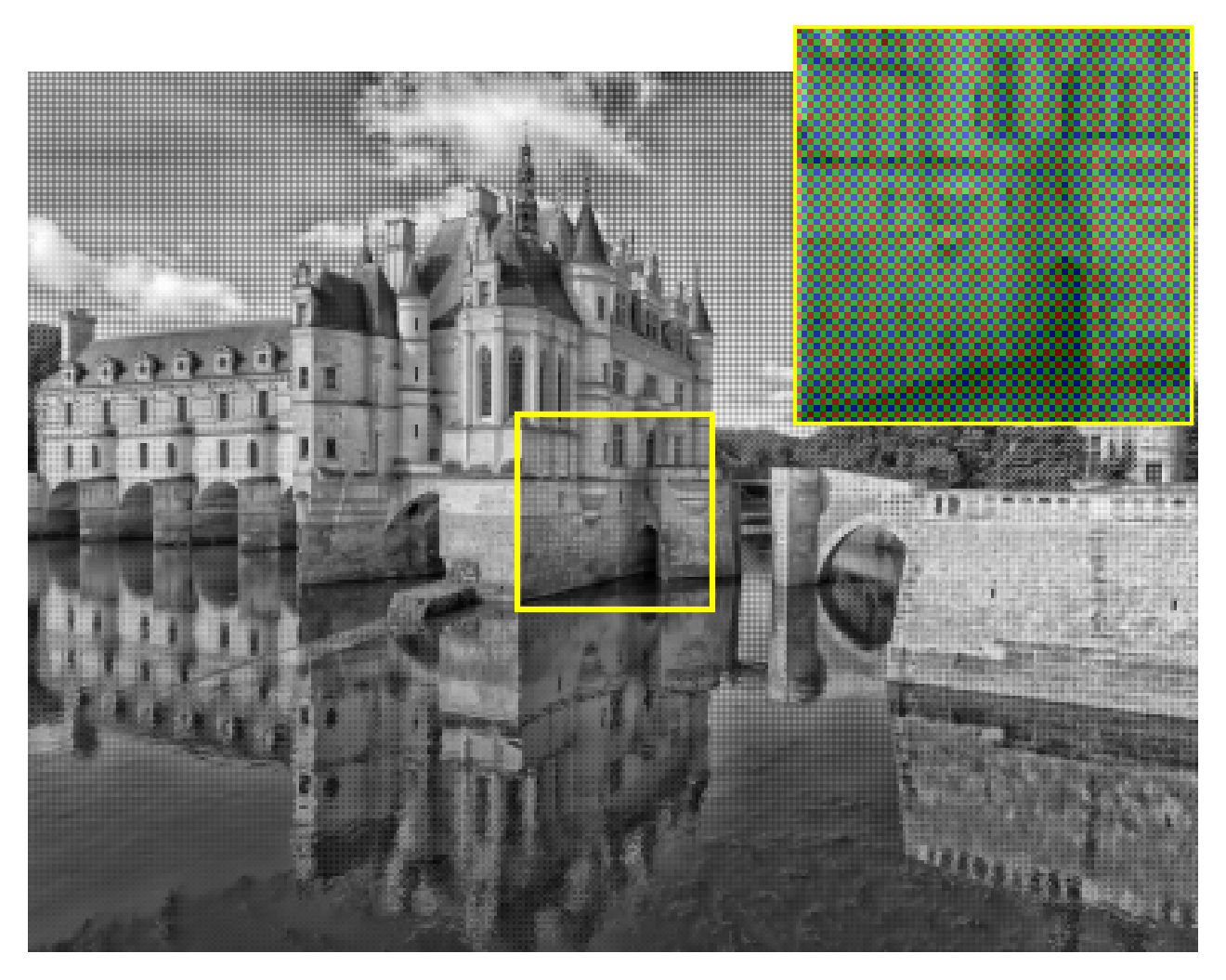}}\hfill
    \subfloat[Demosaic RGB]{\includegraphics[width=0.33\linewidth]{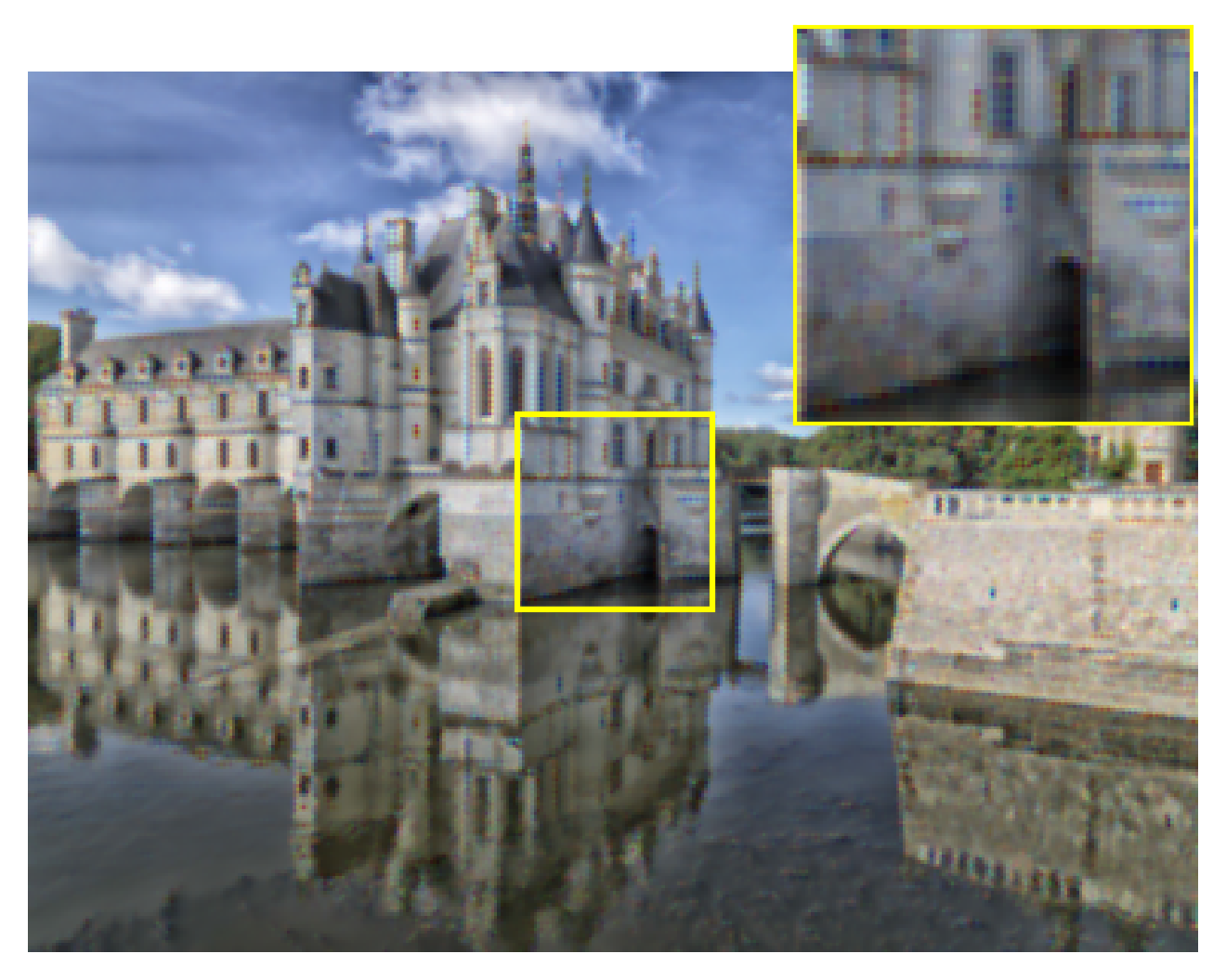}}\hfill
    \caption{Illustration of the CFA pipeline: \textbf{(a)} original RGB scene; \textbf{(b)} RAW CFA mosaic (single‑channel), shown in grayscale with a zoomed‑in patch where a color‑coded RGGB Bayer pattern is overlaid to visualize the sampling locations of each channel rather than true RGB values; and \textbf{(c)} demosaiced RGB reconstruction via simple bilinear interpolation for visualization purpose.}
    \label{fig:cfa_bayer}
    \vspace{-0.3cm}
\end{figure}

\section{Proposed Method: DCCT}
In this section, we introduce the proposed DCCT and present a theoretical analysis to elucidate its behavior.

\subsection{Preliminaries}

Modern digital cameras typically use a single monochrome sensor covered by a CFA so that each photosite measures only one color component~\cite{bayer1976color,lukac2005color,ramanath2005color}.  
The dominant design is the Bayer CFA, a $2\times2$ periodic mosaic with two green, one red, and one blue filter arranged on interleaved quincunx lattices (see Fig.~\ref{fig:cfa_bayer}b)~\cite{bayer1976color,gunturk2005demosaicking}.

Let $\bm I_c(i,j)$ denote the color image for channel $c\in\{\mathrm{R,G,B}\}$ and $\bm M_c(i,j)\in\{0,1\}$ the CFA mask indicating whether channel $c$ is sampled at pixel $(i,j)$. The RAW sensor output can be written as
\begin{equation}\label{eq:mosacing}
\bm Z(i,j)=\sum_{c\in\{\mathrm{R,G,B}\}} \bm M_c(i,j)\, \bm I_c(i,j) + \bm n(i,j),
\end{equation}
where $\bm n$ collects sensor and readout noise~\cite{foi2008practical, yao2021signal}. Thus $\bm Z$ is a single–channel mosaicked image whose nonzero entries follow the CFA’s periodic pattern. To obtain a full RGB image $\hat{\bm I}$, demosaicing reconstructs the missing two components at each location by interpolating neighboring CFA samples according to the known pattern (\eg, Bayer RGGB). In a generic local linear model: 
\begin{equation}~\label{eq:demosaicing}
\hat{\bm I}_c(i,j)=\sum_{(u,v)\in\mathcal{\bm N}(i,j)} \bm h_{c}(u,v;\bm P)\, \bm Z(i+u,j+v),
\end{equation}
where $\mathcal{\bm N}(i,j)$ is a neighborhood, $\bm P$ denotes the CFA pattern, and the interpolation kernel $\bm h_c(\cdot;\bm P)$ depends on both the color channel and the CFA configuration~\cite{zhang2005color,hirakawa2005adaptive, popescu2005exposing,kokkinos2018deep}.

\begin{figure}[]
    \centering
    \includegraphics[page=2,trim=100mm 140mm 90mm 0mm,clip, width=0.495 \textwidth]{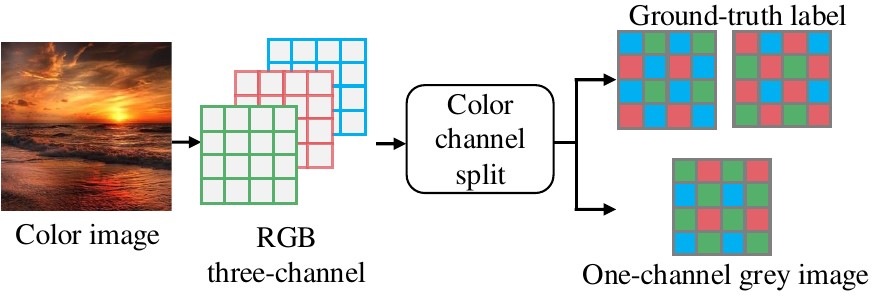}
    \caption{Bayer-like masking strategy, where a single channel (representing the CFA raw input) is retained as the network input, while the remaining two color components serve as the reconstruction targets (ground-truth labels).
    }
    \label{fig_pretext_task}
    \vspace{-0.33cm}
\end{figure}

In the frequency domain, CFA sampling acts as a modulation, \eg, green-channel sampling in a Bayer pattern can be modeled by multiplying the continuous scene $I_{\text{scene}}(i,j)$ with
$M_G(i,j) = \tfrac{1}{2}[1 + \cos(\pi i + \pi j)]$, yielding
\begin{align}
&\fourier{I}_{G,\text{RAW}}(\omega_i, \omega_j)\nonumber \\
&= \fourier{I}_{\text{scene}}(\omega_i, \omega_j) * \left[ \frac{1}{2}\delta(\omega_i, \omega_j) + \frac{1}{2}\delta(\omega_i - \pi, \omega_j - \pi) \right] \nonumber \\
&= \underbrace{\frac{1}{2}\fourier{I}_{\text{scene}}(\omega_i, \omega_j)}_{\text{Baseband Signal}} + \underbrace{\frac{1}{2}\fourier{I}_{\text{scene}}(\omega_i - \pi, \omega_j - \pi)}_{\text{Aliased Signal}}
\label{eq:cfa_spectrum}
\end{align}
Thus, the RAW spectrum of Eq.~\eqref{eq:cfa_spectrum} is a superposition of the original scene spectrum and a shifted copy concentrated near the Nyquist corner $(\pi,\pi)$.

\begin{figure*}[]
    \centering
    \includegraphics[width=\textwidth]{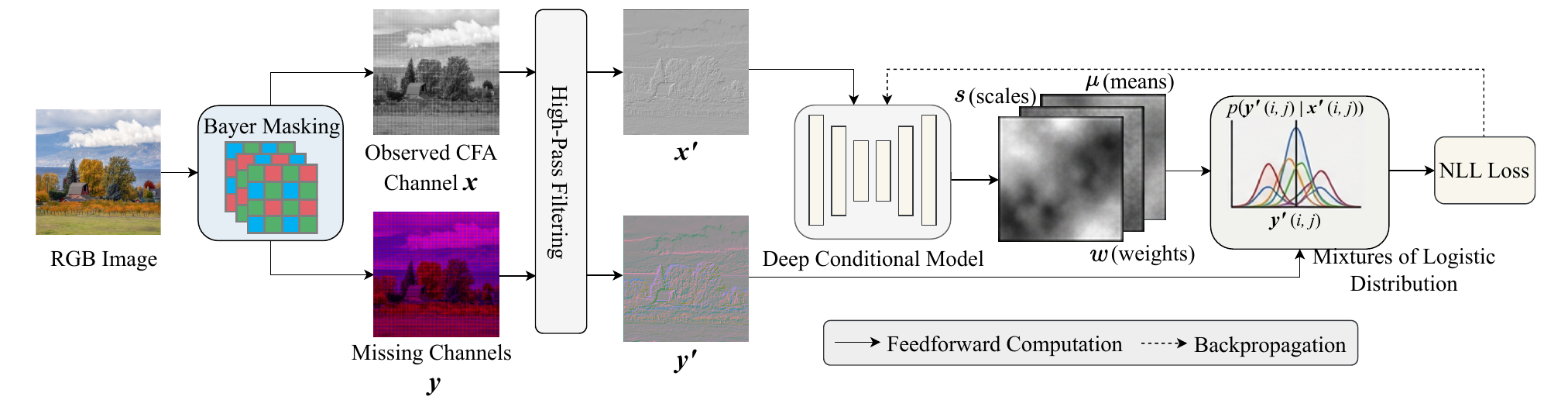}
    \caption{The system diagram of the DCCT via self-supervised reconstruction. }
    \label{fig_estimation_netwrok}
    \vspace{-0.3cm}
\end{figure*}

\subsection{DCCT via Self-Supervised Reconstruction}
Let $C \in \{\texttt{photographic}$, $\texttt{AI-generated}\}$ denote the source class.
Adhering to Bayer CFA geometry, we decompose each local RGB patch into a spatially aligned single-channel observation $\bm x$ and a two-channel residual $\bm y$ using a fixed mask that, at each pixel, retains the single channel value dictated by the $2 \times 2$ Bayer tile for $\bm x$ while assigning the remaining two channels to $\bm y$ (see Fig.~\ref{fig_pretext_task}). 
Thus, $(\bm x,\bm y)$ mimics the class conditional relationship between RAW mosaicked measurements and the missing color samples to be reconstructed during demosaicing, \ie, $p(\bm y \vert \bm x)$.
For photographic images, $p(\bm y \vert \bm x, C{=}\texttt{photographic})$ reflects the local color interpolation behavior induced by CFA sampling, demosaicing, and the ISP pipeline, whereas for AI-generated images, $p(\bm y \vert \bm x, C{=}\texttt{AI-generated})$ arises from a purely digital generation process that directly synthesizes three-channel RGB values. 
Motivated by Eq.~\eqref{eq:cfa_spectrum}, we analyze these color correlations in the high-pass domain, transforming $\bm x$ and $\bm y$ into $\bm x'$ and $\bm y'$ via high-pass filtering.

We parameterize the conditional density $p_{\bm \theta}(\bm y' \vert \bm x')$ using a deep conditional model with a mixture of logistic output layer~\cite{salimans2017pixelcnn}:
\begin{equation}
p_{\bm \theta}(\bm y' \vert \bm x')
= \sum_{k=1}^{K} w_k(x')\,\mathrm{\mathtt{logistic}}\big(\bm y' \vert \mu_k(\bm x'), s_k(\bm x')\big),
\label{eq:mixture_model}
\end{equation}
where $\{w_k,\mu_k,s_k\}_{k=1}^K$ are model outputs and $\mathrm{\mathtt{logistic}}(\cdot)$ denotes the discretized logistic component. The model is trained by minimizing the negative log-likelihood
\begin{equation}
\ell_{\mathrm{NLL}}(\bm \theta)
= - \mathbb{E}_{(\bm x',\bm y')\sim P_{\texttt{photographic}}}
\big[\log p_{\bm \theta}(\bm y' \vert \bm x')\big].
\label{eq:nll_training}
\end{equation}
In practice, due to the sparsity of the high-pass residuals, where the majority of elements cluster around zero despite exhibiting long-tail outliers, we implement value truncation to confine $\bm x'$ and $\bm y'$ within the interval $[-t, t]$.

\begin{table*}[]
\small
\caption{Cross-generator detection accuracy (\%) on the GenImage dataset using SDv1.4 for training, following the protocol of~\cite{zhu2023genimage}. The best results are indicated in bold.}
 \centering
 \resizebox{\textwidth}{!}{
\begin{tabular}{lcccccccccc} 
\toprule
Method     & Venue & Midjourney & SDv1.4 & SDv1.5 & ADM   & GLIDE & Wukong & VQDM  & BigGAN & Avg.  \\
\midrule
CNNSpot~\cite{wang2020cnn}  & CVPR'20  & 84.92      & 99.88  & 99.76  & 53.48 & 53.80 & 99.68  & 55.50 & 49.93  & 74.62 \\
GramNet~\cite{liu2020global}  & CVPR'20  & 73.68      & 98.85  & 98.79  & 51.52 & 55.38 & 95.38  & 55.15 & 49.41  & 72.27 \\
F3Net~\cite{qian2020thinking}   & ECCV'20   & 77.85      & 98.99  & 99.08  & 51.20 & 54.87 & 97.92  & 58.99 & 49.21  & 73.51 \\
CLIP/RN50~\cite{radford2021learning} & ICML'21 & 83.30      & 99.97  & 99.89  & 54.55 & 57.37 & 99.52  & 57.90 & 50.00  & 75.31 \\
Conv-B~\cite{liu2022convnet}  & CVPR'22   & 83.55      & \textbf{99.99}  & \textbf{99.92}  & 51.75 & 56.27 & 99.92  & 58.41 & 50.00  & 74.98 \\
LNP~\cite{liu2022detecting}   &  ECCV'22   & 60.30      & 99.72  & 99.64  & 49.86 & 49.88 & 99.52  & 49.85 & 49.88  & 69.80 \\
DE-FAKE~\cite{sha2022fake} & CCS'23    & 79.88      & 98.65  & 98.62  & 71.57 & 78.05 & 98.42  & 78.31 & 74.37  & 84.73 \\
UnivFD~\cite{ojha2023towards} & CVPR'23   & 91.46      & 96.41  & 96.14  & 58.07 & 73.40 & 94.53  & 67.83 & 57.72  & 79.45 \\
DIRE~\cite{wang2023dire}  &  ICCV'23   & 50.40      & \textbf{99.99}  & \textbf{99.92}  & 52.32 & 67.23 & \textbf{99.98}  & 50.10 & 49.99  & 71.24 \\
LGrad~\cite{tan2023learning}  & CVPR'23    & 84.40      & 99.21  & 99.09  & 59.23 & 83.86 & 98.19  & 57.23 & 61.63  & 80.40 \\
NPR~\cite{tan2024rethinking}   & CVPR'24      & 80.94      & 99.70  & 99.51  & 60.27 & 77.00 & 98.41  & 54.53 & 63.03  & 79.20 \\
DRCT~\cite{chendrct} &  ICML'24    & 91.50      & 95.01  & 94.41  & 79.42 & 89.18 & 94.67  & 90.03 & 81.67  & 89.49 \\
AIDE~\cite{yan2025sanity} & ICLR'25 & 79.38 & 99.74 & 99.76 & 78.54 & 91.82 & 98.65 & 80.26 & 66.89 & 86.88 \\
Effort~\cite{yan2025orthogonal} & ICML'25  & 82.40 & 99.80 & 99.80 & 78.70 & 93.30 & 97.40 & 91.70 & 77.60 & 91.10 \\
\midrule
% DCCT$^\dagger$ (Ours) & -- & 95.16 & 84.11 & 83.99 & 81.65 & 93.19 & 86.89 & 87.37 & 94.48 & 88.36 \\
DCCT (Ours) & -- & \textbf{95.61}      & 99.53  & 99.48  & \textbf{94.02} & \textbf{98.31} & 99.51  & \textbf{98.88} & \textbf{92.84}  & \textbf{97.25} \\
\bottomrule 
\end{tabular}
}
\label{tab_genimage}
\vspace{-0.3cm}
\end{table*}

\subsection{Theoretical Insight}\label{subsec: theorem}
Our analysis centers on a single question fundamental to DCCT: \textit{Does the CFA-aligned high-pass single-to-multi-channel prediction task induce a stable conditional distribution gap between photographic and AI-generated images?}

\begin{proposition}
\label{thm:main}
Let $\bm x'$ and $\bm y'$ denote the high-frequency input and output residuals, respectively. Assume locally Gaussian statistics for the joint distribution of $(\bm x', \bm y')$ within small patches.
There exists a constant $\delta > 0$ such that, for any input $\bm x'$ possessing non-trivial energy $c_x$ on the aliasing subbands $\Omega^\star$, the 1-Wasserstein distance between the high-pass conditional distributions of photographic ($p$) and AI-generated ($q$) images satisfies:
\begin{equation}\label{eq:theorem}
W_1\Big(
p(\bm y'\vert \bm{x}'),
q(\bm y'\vert \bm{x}')
\Big)
\;\ge\; \delta,
\end{equation}
provided that the generative model fails to perfectly replicate the spectral correlations induced by CFA sampling.
\end{proposition}

\noindent\textbf{Heuristic Derivation}.
We begin with the Kantorovich-Rubinstein duality:
\begin{equation}
W_1(p, q) = \sup_{\varphi:\,\mathrm{Lip}(\varphi)\le 1} \left( \E_{\bm{y}' \sim p}[\varphi(\bm{y}')] - \E_{\bm{y}' \sim q}[\varphi(\bm{y}')] \right).
\end{equation}
Let $\bm{\mu}_p(\bm x') := \E_p[\bm{y}' | \bm{x}']$ and $\bm{\mu}_q(\bm x') := \E_q[\bm{y}' | \bm{x}']$.
By selecting the specific 1-Lipschitz test function $\varphi(\bm {y}') = \langle \bm {y}', \bm {v} \rangle$ with $\mathbf{v} = (\boldsymbol{\mu}_p - \boldsymbol{\mu}_q) / \|\boldsymbol{\mu}_p - \boldsymbol{\mu}_q\|_2$, we obtain the lower bound based on the Euclidean distance of means:
\begin{align}~\label{eq:w1_mean_diff_bound}
W_1(p,q)
&\ge 
\langle \bm v, \bm{\mu}_p - \bm{\mu}_q \rangle = \|\bm{\mu}_p(\bm x') - \bm{\mu}_q(\bm x')\|_2.
\end{align}

Starting from Eq.~\eqref{eq:mosacing} and Eq.~\eqref{eq:demosaicing}, $\bm x'$ and $\bm y'$ can be written as a linear transform of the underlying RAW measurements $\bm Z$, denoted as $\bm x'=\mathcal{A}_x \bm Z$ and $\bm y'=\mathcal{A}_y \bm Z$, respectively, governing by the physical CFA sampling and demosaicing pipeline.
Restricting attention to small, locally stationary patches, we adopt the standard Gaussian scale mixture model for natural image statistics~\cite{portilla2003image}, treating $\bm Z$ as Gaussian and thereby implying that the joint distribution of $(\bm x', \bm y')$ is locally Gaussian. The conditional expectation is thus an affine function~\cite{bishop2006pattern}:
\begin{align}
\bm \mu_p(\bm x') = \bm \Sigma_{yx} \bm \Sigma_{xx}^{-1} (\bm x' - \bm \mu_x) + \bm \mu_y.
\end{align}
For AI-generated images, although the global generator is non-linear, we employ a first-order Taylor approximation (local linearization) of the conditional manifold around $\bm x'$. Alternatively, assuming the AI outputs also follow a local Gaussian model (albeit with different covariance structure), we write $\bm \mu_q(\bm x') \approx \bm T_{\mathrm{Gen}} (\bm x' - \bm \mu_x) + \bm \mu_y$, where $\bm T_{\mathrm{Gen}}$ represents the local effective linear mapping of the generator.
Since $\bm x'$ and $\bm y'$ are high-pass residuals, their unconditional means vanish ($\bm \mu_x \approx \bm 0, \bm \mu_y \approx \bm 0$). Thus, we simplify to linear forms $\bm \mu_p(\bm x') = \bm T_{\mathrm{CFA}} \bm x'$ and $\bm \mu_q(\bm x') = \bm T_{\mathrm{Gen}} \bm x'$, where $\bm T_{\mathrm{CFA}} := \bm \Sigma_{yx} \bm \Sigma_{xx}^{-1}$ encodes the specific correlation structure from demosaicing.

Since CFA sampling is linear periodically shift-varying (LPSV) on the pixel grid, a frequency-wise matrix multiplication is not directly available. We therefore move to the $2\times2$ polyphase vector fields~\cite{vetterli1995wavelets} on the coarser lattice, where the same LPSV process becomes a multi-channel linear shift-invariant (LSI) system. Hence, we can write $\widehat{\bm \mu_(\bm x')}(\omega)=\bm T(\omega)\,\widehat{\bm x'}(\omega)$ in the frequency domain. With this multiplicative form and Parseval's theorem (for finite $N$ samples on the coarse grid), we relate the discrepancies in conditional means into integrals over the aliasing band $\Omega^\star$:
\begin{equation}\label{eq:conditinal_mean_gap}
    \|\bm \mu_p - \bm \mu_q\|_2^2 
    = \frac{1}{N} \int_{\Omega^\star} 
        \big\|\big(\bm T_{\mathrm{CFA}}(\omega)-\bm T_{\mathrm{Gen}}(\omega)\big)\widehat{\bm x'}(\omega)\big\|_2^2 \, d\omega,
\end{equation}
Crucially, $\bm T_{\mathrm{CFA}}$ embodies specific aliasing artifacts (spectral coupling) inherent to the CFA, whereas $\bm T_{\mathrm{Gen}}$ arises from a neural network typically optimized in RGB space without such physical constraints. 
We posit that this gap makes the difference between $\bm T_{\mathrm{CFA}}(\omega)$ and $\bm T_{\mathrm{Gen}}(\omega)$ non-degenerate, \ie, there exists a constant $\gamma>0$ such that
\begin{equation}\label{eq:gamma-gap}
    \sigma_{\min}\big(\bm T_{\mathrm{CFA}}(\omega)-\bm T_{\mathrm{Gen}}(\omega)\big) \;\ge\; \gamma > 0, \quad \forall \omega \in \Omega^\star,
\end{equation}
where $\sigma_{\min}(\cdot)$ denotes the minimum singular value.
Using the property $\| \bm A \bm v \|_2 \ge \sigma_{\min}(\bm A) \|\bm v\|_2$, we have:
\begin{align}
    \|\bm \mu_p - \bm \mu_q\|_2^2 
    &\ge \frac{1}{N} \int_{\Omega^\star} 
         \sigma_{\min}^2\big(\bm \Delta(\omega)\big) \|\widehat{\bm x'}(\omega)\|_2^2 \, d\omega \nonumber\\
    &\ge \frac{\gamma^2}{N}
        \int_{\Omega^\star} \|\widehat{\bm x'}(\omega)\|_2^2 \, d\omega 
    \;\ge\; \frac{\gamma^2 c_x}{N},
\end{align}
where the last step uses the assumption of non-trivial energy $c_x$ on $\Omega^\star$.
Substituting this into Eq.~\eqref{eq:w1_mean_diff_bound}, we conclude $W_1(p, q) \ge \gamma \sqrt{c_x/N} := \delta$.

Building on Proposition~\ref{thm:main}, we derive two implications. \textbf{First}, the proposition establishes a positive lower bound on the 1-Wasserstein distance, guaranteeing a non-vanishing gap between the high-pass conditional laws of photographic and AI-generated images under mild physical assumptions. This affirmatively answers our theoretical question: CFA-aligned high-pass single-to-multi-channel prediction induces a stable and intrinsic distribution gap.
\textbf{Second}, this non-vanishing $W_1$ gap arises from this CFA-based task itself, rather than from any ad hoc feature choice. In other words, DCCT is anchored in a task whose underlying conditional laws for photographic and AI-generated images are provably distinct. Proposition~\ref{thm:main} thus theoretically justifies DCCT’s feasibility for generalizable AIGI detection.

\subsection{AIGI Detection}\label{subsec:aigi_detection}
Guided by Sec.~\ref{subsec: theorem}, we train two independent deep conditional networks to model the distributions $p_{\bm \theta}$ (photographic images) and $q_{\bm \varphi}$ (AI-generated images) defined in Eq.\eqref{eq:theorem}, respectively. Once pretrained, we freeze these networks and utilize them as feature extractors. The resulting feature maps, formed by concatenating the predicted color components from both independent models, serve as input for a binary classifier trained with a standard cross-entropy loss~\cite{tan2023learning, wang2023dire}. 
During inference, the input image is processed by these frozen conditional models to generate feature maps, which the classifier uses to predict if the image is photographic or AI-generated.

\begin{table*}[]
\setlength\tabcolsep{2.5pt}
 \centering
 \caption{Cross-generator detection accuracy (\%) on the DRCT-2M dataset using SDv1.4 for training, following the protocol of~\cite{chendrct}. The best results are indicated in bold.}
\resizebox{\textwidth}{!}{
\begin{tabular}{lccccccccccccccccc}
\toprule
\multirow{3}{*}{Method} & \multicolumn{6}{c}{SD Variants}                                                                   & \multicolumn{2}{c}{Turbo Variants}                                                                          & \multicolumn{2}{c}{LCM Variants}                                                                            & \multicolumn{3}{c}{ControlNet Variants}                                                                                                                            & \multicolumn{3}{c}{DR Variants}                                                                                                                              & \multirow{3}{*}{Avg.}  \\
\cmidrule(lr){2-7} \cmidrule(lr){8-9} \cmidrule(lr){10-11} \cmidrule(lr){12-14} \cmidrule(lr){15-17}
                        & LDM   & SDv1.4 & SDv1.5 & SDv2  & SDXL  & \begin{tabular}[c]{@{}c@{}}SDXL-\\ Refiner\end{tabular} & \begin{tabular}[c]{@{}c@{}}SD-\\ Turbo\end{tabular} & \begin{tabular}[c]{@{}c@{}}SDXL-\\ Turbo\end{tabular} & \begin{tabular}[c]{@{}c@{}}LCM-\\ SDv1.5\end{tabular} & \begin{tabular}[c]{@{}c@{}}LCM-\\ SDXL\end{tabular} & \begin{tabular}[c]{@{}c@{}}SDv1-\\ Ctrl\end{tabular} & \begin{tabular}[c]{@{}c@{}}SDv2-\\ Ctrl\end{tabular} & \begin{tabular}[c]{@{}c@{}}SDXL-\\ Ctrl\end{tabular} & \begin{tabular}[c]{@{}c@{}}SDv1-\\ DR\end{tabular} & \begin{tabular}[c]{@{}c@{}}SDv2-\\ DR\end{tabular} & \begin{tabular}[c]{@{}c@{}}SDXL-\\ DR\end{tabular} \\
                        \midrule
CNNSpot & 99.87 & 99.91  & 99.90  & 97.55 & 66.25 & 86.55 & 86.15 & 72.42 & 98.26 & 61.72 & 97.96 & 85.89 & 82.84 & 60.93 & 51.41 & 50.28 & 81.12 \\
GramNet                 & 99.40 & 99.01  & 98.84  & 95.30 & 62.63 & 80.68                                                   & 71.19                                               & 69.32                                                 & 93.05                                                 & 57.02                                               & 89.97                                                & 75.55                                                & 82.68                                                & 51.23                                              & 50.01                                              & 50.08                                              & 76.62 \\
F3Net                   & 99.85 & 99.78  & 99.79  & 88.66 & 55.85 & 87.37                                                   & 68.29                                               & 63.66                                                 & 97.39                                                 & 54.98                                               & 97.98                                                & 72.39                                                & 81.99                                                & 65.42                                              & 50.39                                              & 50.27                                              & 77.13 \\
CLIP/RN50               & 99.00 & 99.99  & 99.96  & 94.61 & 62.08 & 91.43                                                   & 83.57                                               & 64.40                                                 & 98.97                                                 & 57.43                                               & 99.74                                                & 80.69                                                & 82.03                                                & 65.83                                              & 50.67                                              & 50.47                                             & 80.05 \\
Conv-B                  & \textbf{99.97} & \textbf{100.00} & 99.97  & 95.84 & 64.44 & 82.00                                                   & 80.82                                               & 60.75                                                 & 99.27                                                 & 62.33                                               & 99.80                                                & 83.40                                                & 73.28                                                & 61.65                                              & 51.79                                              & 50.41                                              & 79.11 \\
LNP                     & 49.50 & 99.45  & 99.45  & 51.26 & 56.37 & 71.24                                                   & 99.32                                               & 99.44                                               & 98.96                                                 & 65.30                                               & 66.07                                                & 71.02                                                & 69.81                                                & 50.71                                              & 50.65                                              & 49.82                                              & 71.77 \\
DE-FAKE                 & 92.10 & 99.53  & 99.51  & 89.65 & 64.02 & 69.24                                                   & 92.00                                               & 93.93                                                 & 99.13                                                 & 70.89                                               & 58.98                                                & 62.34                                                & 66.66                                                & 50.12                                              & 50.16                                              & 50.00                                              & 75.52 \\
UnivFD                  & 98.30 & 96.22  & 96.33  & 93.83 & 91.01 & 93.91                                                   & 86.38                                               & 85.92                                                 & 90.44                                                 & 88.99                                               & 90.41                                                & 81.06                                                & 89.06                                                & 51.96                                              & 51.03                                              & 50.46                                              & 83.46 \\
DIRE                    & 98.19 & 99.94  & 99.96  & 68.16 & 53.84 & 71.93                                                   & 58.87                                               & 54.35                                                 & 99.78                                                 & 59.73                                               & 99.65                                                & 64.20                                                & 59.13                                                & 51.99                                              & 50.04                                              & 49.97                                              & 71.23 \\
LGrad                   & 96.79 & 97.87  & 97.93  & 87.80 & 54.52 & 77.07                                                   & 98.11                                               & 98.20                                                & 97.80                                                 & 59.75                                               & 94.58                                                & 92.93                                                & 90.67                                                & 49.44                                              & 49.36                                              & 49.44                                              & 80.70 \\
NPR                     & 94.92 & 97.73  & 97.78  & 82.96 & 67.52 & 79.72                                                   & 95.81                                               & 95.85                                                 & 97.68                                                 & 80.32                                               & 94.45                                                & 92.96                                                & 90.22                                                & 49.24                                              & 49.41                                              & 49.01                                              & 83.63 \\
DRCT                   & 96.74 & 96.26  & 96.33  & 94.89 & 96.24 & 93.46                                                   & 93.43                                               & 92.94                                                 & 91.17                                                 & 95.01                                               & 95.60                                                & 92.68                                                & 91.95                                                & \textbf{94.10}                                              & \textbf{69.55}                                              & \textbf{57.43}                                              & 90.49 \\
AIDE & 94.05 & 96.77 & 96.71 & 86.94 & 65.76 & 80.48 & 77.54 & 73.84 & 97.95 & 72.88 & 89.07 & 72.29 & 75.03 & 49.96 & 49.95 & 49.95 & 76.20 \\
Effort & 99.95 & \textbf{100.00} & \textbf{100.00} & \textbf{99.93} & \textbf{99.74} & \textbf{99.73} & \textbf{99.85} & \textbf{99.75} & \textbf{99.89} & 99.53 & \textbf{99.97} & 98.99 & \textbf{99.76} & 50.00 & 50.00 & 50.00 & 90.44 \\
\midrule
DCCT (Ours)       & 99.70 & 99.70  & 99.70  & 99.32 & 99.70 & 99.70       & 99.70          & 99.70            & 99.70            & \textbf{99.70}         & 99.39       & \textbf{99.14}      & 99.47      & 54.10     & 54.51    & 54.95    &        \textbf{91.13}        \\
\bottomrule
\end{tabular}
}
 \label{tab_diffusion_2m}
 \vspace{-.3cm}
\end{table*}

\section{Experiments}
\subsection{Experimental Setups }
\noindent\textbf{Datasets.}
We evaluate the DCCT on two widely used benchmarks: GenImage~\cite{zhu2023genimage} and DRCT-2M~\cite{chendrct}. GenImage contains images synthesized by 8 generative models--Midjourney~\cite{midjourney}, Stable Diffusion v1.4 (SDV1.4)\cite{rombach2022high}, SDV1.5, ADM\cite{dhariwal2021diffusion}, GLIDE~\cite{nichol2021glide}, Wukong~\cite{wukong}, VQDM~\cite{gu2022vector}, and BigGAN~\cite{brock2018large}---each paired with photographic counterparts from ImageNet~\cite{deng2009imagenet}.
DRCT-2M focuses on diffusion-based synthesis, covering 16 Stable Diffusion models and variants, including LDM~\cite{rombach2022high}, SDv2, SDXL~\cite{podell2023sdxl}, and their turbo variants, latent consistency model (LCM) variants~\cite{luo2023lcm}, ControlNet variants~\cite{zhang2023adding}, and diffusion reconstruction (DR) variants, all paired with photographs from MSCOCO~\cite{lin2014microsoft}.  

\noindent\textbf{Implementation Details.} 
We instantiate the deep conditional network using a U-Net~\cite{ronneberger2015u}, configuring the logistic output with $K=10$ mixture components (Eq.\eqref{eq:mixture_model}) and setting the residual truncation threshold to $t=7$. Following prior work~\cite{fridrich2012rich, zhong2023patchcraft}, we utilize 30 high-pass filters from Fridrich and Kodovsk'{y} for high-pass transformation.
For the classifier described in Sec.\ref{subsec:aigi_detection}, we employ a shallow ResNet~\cite{he2016deep} with four residual blocks, followed by a 2-layer Transformer encoder~\cite{dosovitskiy2020image} and a final fully-connected layer. 
Both the deep conditional network and the classifier are optimized via Adam~\cite{kingma2014adam} with a learning rate of $10^{-4}$ and a batch size of 16. To improve the generalization~\cite{ojha2023towards, wang2020cnn}, we apply JPEG compression as a benign perturbation during training, with the quality factor (QF) sampled from $\mathcal{U}(70, 100)$ and applied with 5\% probability.
We train on random $64\times64\times3$ crops and compute the final test score by averaging predictions across 16 patches per image.

\begin{table*}[]
\small
\centering
\caption{Generalization (Acc) of AI-generated image detectors to emerging generators.}
% \resizebox{\linewidth}{!}{
\begin{tabular}{lcccccccccc}
\toprule
Method          & GigaGAN & DFGAN  & GALIP & FLUX.1 & FLUX.1-Kontext & SD-3.5-Turbo & Qwen-Image &  Avg. \\
\midrule
UnivFD          & 33.20 & 21.50  & 35.40 & 76.40 & 64.75 & 75.80 & 33.60 & 48.66  \\
Effort          & \textbf{99.60} &  \textbf{99.70}  & 72.00 & 93.44 & 91.65 & 95.37 & 96.93 &  92.67 \\
\midrule
DCCT (Ours)     & 99.40 & 98.90 & \textbf{100.00} & \textbf{99.10} & \textbf{98.10} & \textbf{96.20} & \textbf{99.86} & \textbf{98.79}  \\
\bottomrule
\end{tabular}
% }
\label{tab: more_gans_in_the_wild}
\end{table*}

\begin{figure*}[ht]
    \centering
    \includegraphics[width=\textwidth]{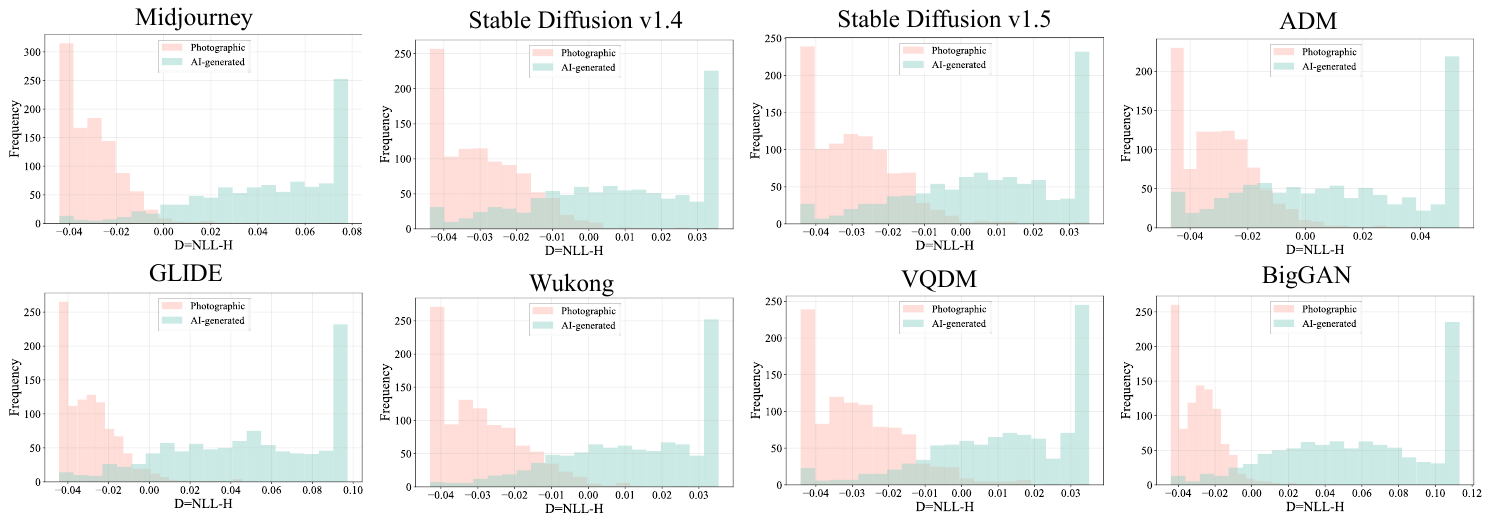}
    \caption{ Distributions of the anomaly score $D$ for photographic \textit{vs}. AI-generated images on the GenImage dataset~\cite{zhu2023genimage}. We compute $D$ by averaging the pixel-wise difference between the negative log-likelihood $\mathrm{NLL}$ and the entropy $H$ of the color feature maps predicted by DCCT (trained exclusively on photographic images). The resulting distributions demonstrate a clear separation between photographic and AI-generated content across various generators.}
    \label{fig_d_distribution}
    \vspace{-.3cm}
\end{figure*}

\subsection{Main Results}
To assess generalizability, we test the proposed detector on GenImage and DRCT-2M and report detection accuracy (\%). Consistent with \cite{zhu2017unpaired, chendrct}, training is performed using SDv1.4-generated images and photographic counterparts.  

\noindent\textbf{GenImage.} 
As shown in Table~\ref{tab_genimage}, while all detectors achieve near-perfect accuracy when the test data, \ie, SDv1.4 and SDv1.5, match the training source, baseline performance deteriorates sharply on mismatched generators such as BigGAN~\cite{brock2018large} and ADM~\cite{dhariwal2021diffusion}. In contrast, our method maintains over 97\% accuracy across these generators and outperforms the state-of-the-art Effort~\cite{yan2025orthogonal} by an average of 8\%. 
This substantial gain arises from modeling color correlations introduced by the in-camera imaging pipeline, rather than relying on transient generator-specific artifacts~\cite{liu2020global, qian2020thinking, sha2022fake, wang2023dire, tan2024rethinking, yan2025sanity}. By directly exploiting these generator-agnostic color correlation patterns, our approach avoids overfitting and generalizes robustly across diverse generative models.
Beyond detectors trained from scratch, Table~\ref{tab_genimage} also shows that our pretrained features outperform generator-agnostic pretraining schemes such as LNP~\cite{liu2022detecting}, UnivFD~\cite{ojha2023towards}, and CLIP/RN50~\cite{radford2021learning}. UnivFD and CLIP/RN50 adopt CLIP-style semantic pretraining, while LNP uses denoising-based pretraining, but none of them explicitly model in-camera imaging characteristics. In contrast, DCCT is tailored to these camera-induced statistics, yielding substantially stronger performance.

\noindent\textbf{DRCT-2M.} 
Table~\ref{tab_diffusion_2m} reports the results on DRCT-2M. Our method achieves nearly 100\% detection accuracy across most generative models, with the only notable failures on the three Diffusion Reconstruction (DR) variants. In these DR images, only a small region is synthesized while most pixels remain unaltered photographic content. Consequently, their global statistics, and thus their color correlations, closely match those of photographic images, making them intrinsically difficult to distinguish as AI-generated. Most baselines likewise struggle on these DR subsets, with the exception of DRCT, whose training procedure explicitly uses DR images as hard pseudo-AI-generated samples.
Beyond absolute performance, Table~\ref{tab_diffusion_2m} also shows that artifact-based detectors, \eg, F3Net, DE-FAKE, DIRE, LGrad, and NPR, generalize poorly even within the Stable Diffusion family: when trained on SDv1.4, many methods degrade notably on other SD variants (SDXL, Turbo, LCM, and ControlNet variants). This instability indicates a strong reliance on generative artifacts, whereas our camera-driven modeling yields much more reliable cross-variant generalization.

\subsection{Further Analysis}\label{subsec:further_analysis}
\noindent\textbf{Feature Separability Analysis.}
To assess the effectiveness of our learned color correlations, we study a one-class anomaly detection variant, namely DCCT$^\dagger$, where we use the pixel-wise difference between the negative log-likelihood and the entropy~\cite{cozzolino2024zero} of the predicted color distribution:
\begin{equation}
\begin{aligned}
\mathrm{NLL}{i,j}&=-\log p_{\bm \theta}(\bm y_{i,j} \vert \bm x_{i,j}), \\
H_{i,j}&=-\sum_k p_{\bm \theta}(k \vert \bm x_{i,j})\log p_{\bm \theta}(k \vert \bm x_{i,j}),
\end{aligned}
\end{equation}
and average these pixel-wise values over all locations to obtain an anomaly score $D=\frac{1}{\mathcal{\vert I\vert}\mathcal{\vert J\vert}}\sum_{i=1}^{\mathcal{\vert I\vert}}\sum_{j=1}^{\mathcal{\vert J\vert}}(\mathrm{NLL}_{i,j}-H_{i,j})$. We then estimate the score distribution on photographic data, and at test time flag images whose score falls outside the distribution as AI-generated.
As illustrated in Fig.\ref{fig_d_distribution}, we observe a significant divergence in anomaly scores between photographic and AI-generated images. This disparity not only validates the effectiveness of learned color correlations but also suggests the feasibility of identifying AI-generated images using only photographic priors. 
Concretely, we set a threshold at the 95th percentile of the training set's scores and classify samples exceeding this threshold as AI-generated. 
Remarkably, DCCT$^\dagger$ achieves 88.36\% accuracy, surpassing most binary detectors in Table\ref{tab_genimage} despite never having been trained on AI-generated samples.

\begin{figure}
    \centering
    \subfloat[CLIP]{\includegraphics[width=0.45\linewidth]{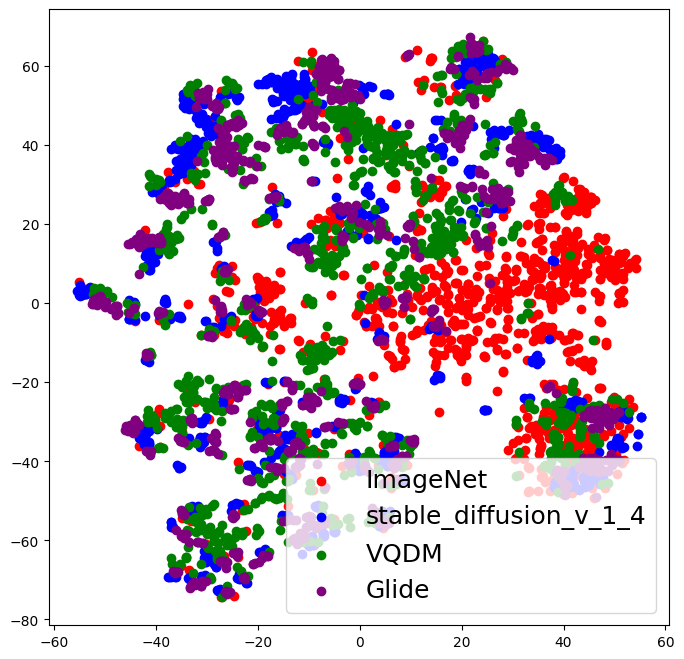}}\hfill
    \subfloat[DCCT]{\includegraphics[width=0.45\linewidth]{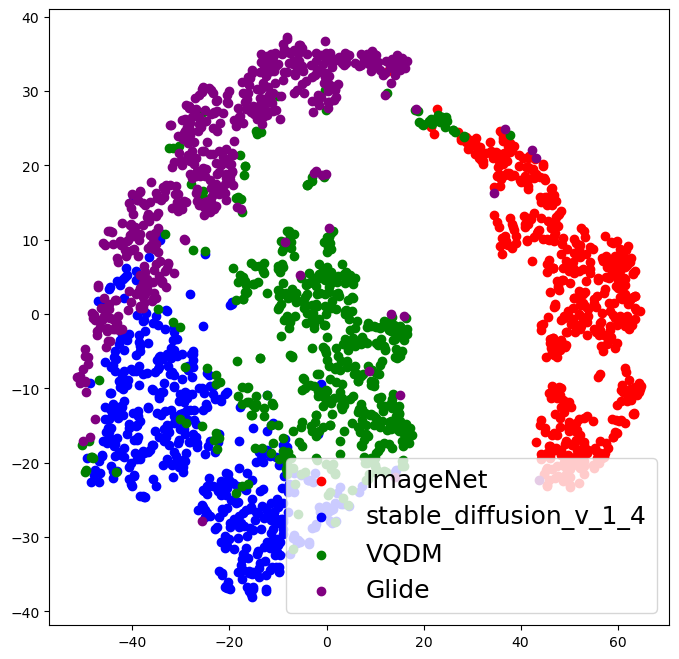}}
    \caption{t-SNE embeddings of learned features for photographic and AI-generated images.}
    \label{fig:tsne}
    \vspace{-.6cm}
\end{figure}

\begin{figure*}[ht]
    \centering
    \includegraphics[width=0.93\linewidth]{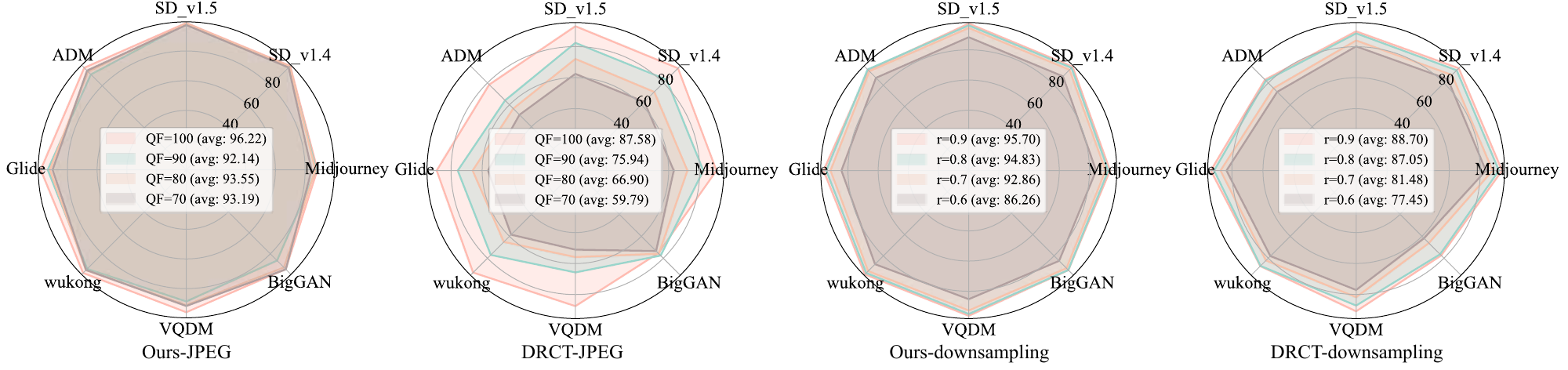}
    \caption{Robustness of detectors to benign post-processing operations. Results show average accuracy across eight GenImage generators~\cite{zhu2023genimage} for the varying intensity levels listed in the legend.
 }
    \label{fig_robust}
    \vspace{-.1cm}
\end{figure*}

We also investigate the feature separability of the standard DCCT. Fig.~\ref{fig:tsne} presents a t-SNE visualization~\cite{van2008visualizing} of the feature embeddings for both photographic and AI-generated images. The plot reveals a clear separation between the two categories, demonstrating the model's high discriminative capability.

%-------------------------------------------------------------------------
\begin{table*}[t]
\small
\centering
\caption{Ablations of key design in DCCT, measured by mean accuracy (mAcc) on GenImage. The default setting is shaded \colorbox{lightgray!20}{gray}.
}
\vspace{-0.2cm}
\captionsetup{font=small}
    \begin{subtable}[t]{0.21\textwidth}
        \centering
        \renewcommand\arraystretch{1.0}
        \caption{Deep conditional models.}
        % \resizebox{\linewidth}{!}{
        \begin{tabular}{cc|c} 
        \toprule
        $p_{\bm \theta}$ & $q_{\bm \varphi}$ & mAcc \\
        \hline
        \cmark & \xmark & 95.48  \\
        \xmark & \cmark & 92.23 \\
        \cellcolor{lightgray!20}{\cmark}   & \cellcolor{lightgray!20}{\cmark} & \cellcolor{lightgray!20}{\textbf{97.27}}\\
        \bottomrule
        \end{tabular}
        % }
        \label{tab:ablation_deep_conditional_model}
    \end{subtable}
    \hfill
    \begin{subtable}[t]{0.32\textwidth}
        \centering
        \renewcommand\arraystretch{1.0}
        \caption{Model structure.}
        \resizebox{\linewidth}{!}{
        \begin{tabular}{cc|c}
         \toprule
        High-pass filters & CFA mask & mAcc \\
        \hline
        \xmark & \cmark & 84.32  \\
        \cmark & \xmark & 89.92  \\
        \cellcolor{lightgray!20}{\cmark}   & \cellcolor{lightgray!20}{\cmark}   & \cellcolor{lightgray!20}{\textbf{97.27}} \\
        \bottomrule
        \end{tabular}
        }
        \label{tab:ablation_model_structure} 
     \end{subtable}
      \hfill
     \begin{subtable}[t]{0.24\textwidth}
        \centering
        \renewcommand\arraystretch{1.0}
        \caption{Truncation threshold.}
        % \resizebox{0.95\linewidth}{!}{
        \begin{tabular}{c|c}
         \toprule
        $t$ value & mAcc \\
        \hline
        3 & 97.21  \\
        15 & 95.18  \\
        \cellcolor{lightgray!20}{7} & \cellcolor{lightgray!20}{\textbf{97.27}} \\
        \bottomrule
        \end{tabular}
        % }
        \label{tab:ablation_t} 
     \end{subtable}
    \hfill
    \begin{subtable}[t]{0.21\textwidth}
        \centering
        \renewcommand\arraystretch{1.0}
        \caption{Training strategy.}
        \resizebox{\linewidth}{!}{
        \begin{tabular}{c|c}
        \toprule
         Full finetuning   & mAcc  \\
        \hline
        \cmark & 93.45  \\
        \cellcolor{lightgray!20}{\xmark} & \cellcolor{lightgray!20}{\textbf{97.27}} \\
        \bottomrule
        \end{tabular}}
        \label{tab:ablation_training_strategy}
    \end{subtable}
\label{tab: ablations_all}
\vspace{-.3cm}
\end{table*}
%-------------------------------------------------------------------------

\noindent\textbf{Robustness Evaluation.}
We further examine the resilience of our detector to benign perturbations common on social platforms, which often erase low-level statistical cues. We test under varying intensities of JPEG compression ($\mathrm{QF} \in [70, 100]$) and spatial downsampling ($r \in [0.6, 0.9]$). Using DRCT~\cite{chendrct} as the primary baseline, Fig.~\ref{fig_robust} illustrates the results. Our detector maintains a significant lead under all perturbation levels, indicating that our learned features are more robust in handling post-processed content.

\noindent\textbf{Generalization to emerging generators.}
We extended our evaluation to rapidly evolving generative models by collecting AI-generated images from GigaGAN~\cite{kang2023scaling}, DFGAN~\cite{tao2022df}, GALIP~\cite{tao2023galip}, FLUX.1~\cite{flux}, FLUX.1-Kontext~\cite{flux}, SD-3.5-Turbo~\cite{sd3_5}, and Qwen-Image~\cite{wu2025qwen}.
As shown in Table~\ref{tab: more_gans_in_the_wild}, DCCT demonstrates exceptional superiority and forward-compatibility, effectively generalizing to unseen architectures.
In contrast, UnivFD degrades sharply in this regime, revealing the limits of semantics-oriented CLIP features. Effort, though significantly stronger than UnivFD, still trails DCCT, highlighting the superiority of our camera-intrinsic representation.

\subsection{Ablation Studies}

\noindent\textbf{Deep Conditional Models.}
We assess the necessity of our dual-network design by restricting the DCCT to a single color correlation learning. As shown in Table \ref{tab:ablation_deep_conditional_model}, performance degrades when either network is removed. These results corroborate our theoretical analysis in Sec~\ref{subsec: theorem}, demonstrating that using two specialized networks provides superior discriminative features than using a single one.

\noindent\textbf{Model Structure.} 
We \textbf{first} evaluate  the role of high-pass filters by removing them, forcing the model to rely solely on spatial inputs. As shown in Table \ref{tab:ablation_model_structure}, performance degrades substantially in this degenerated setting, hightlighting the necessity of the high-frequency domain.  We argue this is expected because modern generators are explicitly optimized for image space distribution matching via adversarial learning~\cite{goodfellow2014generative} or denoising/noise prediction in diffusion~\cite{ho2020denoising}, making spatial statistics (often reflected by FID~\cite{heusel2017gans}) easy to mimic.

We \textbf{then} investigate the benefit of formulating color correlation learning using a Bayer-filter-based mask (see Fig.~\ref{fig_pretext_task}) compared to a random selection of R, G, or B components. As shown in Table \ref{tab:ablation_model_structure}, the random mask leads to overall performance degradation. This finding suggests that aligning the masking strategy with physical imaging signatures is crucial for capturing the subtle inter-channel correlations that distinguish camera output from generated content.

\noindent\textbf{Truncation Threshold.}
We ablate the truncation threshold $t$ by both shrinking and enlarging it (see Table~\ref{tab:ablation_t}). Increasing to $t=15$ harms performance relative to the default $t=7$, likely because the concentrated residual signal is diluted by sparse, non-informative values. The similar performance of of $t=3$ and $t=7$ suggests that the most discriminative information lies within a narrow band around zero.

\noindent\textbf{Training Strategy.}
We investigate the training strategy by unfreezing the two correlation learning networks during the classification head update. As shown in Table \ref{tab:ablation_training_strategy}, full model fine-tuning yields inferior results compared to the frozen baseline. This confirms that the fixed correlation learning modules provide stable, high-quality feature representations, and that restricting updates to the classification head is sufficient for optimal detection performance.

\section{Conclusion and Discussion}
We have introduced DCCT for detecting AI-generated images by exploiting intrinsic traces of the camera imaging pipeline. 
At its core, a demosaicing-guided pretraining stage models the conditional distribution of missing color channels from high-frequency residuals, capturing inter-channel dependencies characteristic of CFA sampling and demosaicing. Extensive experiments demonstrate that our detector achieves state-of-the-art generalization across unseen generators and superior robustness to benign post-processing.
Future work includes improving DCCT on mixed-content images (\eg, DR images) and strengthening robustness against generators that mimic CFA statistics. Beyond Bayer CFAs, extending DCCT to alternative sensor designs, such as non-Bayer CFAs, multi-spectral sensors, or computational photography pipelines, also remains a promising direction.

% \section*{Impact Statement}
% This research advances application-driven machine learning by developing a method for detecting AI-generated images. Through effectively identifying visual misinformation and impersonation, it contributes to safer, more trustworthy digital media. Potential negative impacts include adversaries using our detector to improve future generative models and unequal or discriminatory deployment of detection tools. We therefore recommend controlled access, careful documentation of limitations and biases, and ongoing monitoring and mitigation of misuse, so that the positive societal benefits of this line of work outweigh its risks.

{
    \small
    \bibliographystyle{icml2026}
    \bibliography{DCCT}
}

%%%%%%%%%%%%%%%%%%%%%%%%%%%%%%%%%%%%%%%%%%%%%%%%%%%%%%%%%%%%%%%%%%%%%%%%%%%%%%%
%%%%%%%%%%%%%%%%%%%%%%%%%%%%%%%%%%%%%%%%%%%%%%%%%%%%%%%%%%%%%%%%%%%%%%%%%%%%%%%
% APPENDIX
%%%%%%%%%%%%%%%%%%%%%%%%%%%%%%%%%%%%%%%%%%%%%%%%%%%%%%%%%%%%%%%%%%%%%%%%%%%%%%%
%%%%%%%%%%%%%%%%%%%%%%%%%%%%%%%%%%%%%%%%%%%%%%%%%%%%%%%%%%%%%%%%%%%%%%%%%%%%%%%
\newpage
\appendix
\onecolumn

\begin{table*}[]
\small
\caption{Cross-generator detection accuracy (\%) on the GenImage dataset using SDv1.4 for training, following the protocol of~\cite{zhu2023genimage}. The best two results are indicated in bold.}
 \centering
 \resizebox{\textwidth}{!}{
\begin{tabular}{lcccccccccc} 
\toprule
Method     & Venue & Midjourney & SDv1.4 & SDv1.5 & ADM   & GLIDE & Wukong & VQDM  & BigGAN & Avg.  \\
\midrule
Conv-B~\cite{liu2022convnet}  & CVPR'22   & 83.55      & \textbf{99.99}  & \textbf{99.92}  & 51.75 & 56.27 & \textbf{99.92}  & 58.41 & 50.00  & 74.98 \\
LNP~\cite{liu2022detecting}   &  ECCV'22   & 60.30      & 99.72  & 99.64  & 49.86 & 49.88 & 99.52  & 49.85 & 49.88  & 69.80 \\
DE-FAKE~\cite{sha2022fake} & CCS'23    & 79.88      & 98.65  & 98.62  & 71.57 & 78.05 & 98.42  & 78.31 & 74.37  & 84.73 \\
UnivFD~\cite{ojha2023towards} & CVPR'23   & 91.46      & 96.41  & 96.14  & 58.07 & 73.40 & 94.53  & 67.83 & 57.72  & 79.45 \\
DIRE~\cite{wang2023dire}  &  ICCV'23   & 50.40      & \textbf{99.99}  & \textbf{99.92}  & 52.32 & 67.23 & \textbf{99.98}  & 50.10 & 49.99  & 71.24 \\
LGrad~\cite{tan2023learning}  & CVPR'23    & 84.40      & 99.21  & 99.09  & 59.23 & 83.86 & 98.19  & 57.23 & 61.63  & 80.40 \\
NPR~\cite{tan2024rethinking}   & CVPR'24      & 80.94      & 99.70  & 99.51  & 60.27 & 77.00 & 98.41  & 54.53 & 63.03  & 79.20 \\
DRCT~\cite{chendrct} &  ICML'24    & \textbf{91.50}      & 95.01  & 94.41  & \textbf{79.42} & 89.18 & 94.67  & \textbf{90.03} & \textbf{81.67}  & \textbf{89.49} \\
AIDE~\cite{yan2025sanity} & ICLR'25 & 79.38 & 99.74 & 99.76 & 78.54 & \textbf{91.82} & 98.65 & 80.26 & 66.89 & 86.88 \\
\midrule
DCCT$^\dagger$ (One-class Variant of Ours ) & -- & \textbf{95.16} & 84.11 & 83.99 & \textbf{81.65} & \textbf{93.19} & 86.89 & \textbf{87.37} & \textbf{94.48} & \textbf{88.36} \\
\bottomrule 
\end{tabular}
}
\label{tab_genimage_dcct_oc}
% \vspace{-0.3cm}
\end{table*}

\section{Overall DCCT Pipeline and Algorithm}
This section details the DCCT pipeline, including both the training and inference stages. For clarity and reproducibility, Algorithm~\ref{alg:dcct_train} and Algorithm~\ref {alg:dcct_infer} provide a step-by-step description of the full procedure.  

\begin{algorithm}[t]
\small
\caption{DCCT: Training Phase}
\label{alg:dcct_train}

\renewcommand{\algorithmicrequire}{\textbf{Input:}}
\renewcommand{\algorithmicensure}{\textbf{Output:}}
\begin{algorithmic}[1]
\REQUIRE
Photo dataset $\mathcal{D}_{\text{photo}}$; AI dataset $\mathcal{D}_{\text{AI}}$; labels $c\in\{0,1\}$ for $\mathcal{D}_{\text{photo}}\cup\mathcal{D}_{\text{AI}}$;
CFA masks $\{M_R,M_G,M_B\}$; high-pass filters $\mathcal{\bm H}=\{\bm h_m\}_{m=1}^{M}$; truncation threshold $t$;
Patch size $s$; mixture size $K$.
\ENSURE Trained conditional model $\bm f_{\bm \theta}(\cdot)$, $\bm f_{\bm \phi}(\cdot)$, and classifier $\bm g_{\bm \psi}(\cdot)$.

\vspace{2pt}
\STATE \textit{$\triangleright$} \textbf{Stage I: Training conditional models}
\STATE Initialize $\bm \theta$ and $\bm \phi$
\STATE \textit{$\triangleright$} \textbf{Stage I-A: Training $\bm f_{\bm \theta}(\cdot)$ on $\mathcal{D}_{\text{photo}}$}
\FOR{each epoch}
  \FOR{each mini-batch of images $\bm I \sim \mathcal{D}_{\text{photo}}$}
    \STATE Random crop $\tilde{\bm I}\in\mathbb{R}^{s\times s\times 3}$ from $\bm I$
    \STATE $\bm x \gets \bm M_R\odot \tilde{\bm I}_R + \bm M_G\odot \tilde{\bm I}_G + \bm M_B\odot \tilde{\bm I}_B$;\;\; $y \gets \textsc{PackMissingChannels}(\tilde{I};M_R,M_G,M_B)$
    \STATE $\bm x' \gets \textsc{Truncate}(\textsc{Stack}(\{\bm h_m * \bm x\}_{m=1}^{M}),-t,t)$;\;\; $\bm y' \gets \textsc{Truncate}(\textsc{Stack}(\{\bm h_m * \bm y\}_{m=1}^{M}),-t,t)$
    \STATE Per-pixel mixture parameters $\{\bm w_k,\bm \mu_k,\bm s_k\}_{k=1}^{K} \gets f_\theta(x')$ 
    \STATE $p_{\bm \theta}(\bm y'|\bm x')=\prod_{(i,j)}\sum_{k=1}^{K} \bm w_k(i,j)\cdot \textsc{Logistic}\!\left(\bm y'(i,j)\mid \bm \mu_k(i,j), \bm s_k(i,j)\right)$
    \STATE $\ell_{\mathrm{NLL}}(\bm \theta)\gets -\log p_{\bm \theta}(\bm y'|\bm x')$
    \STATE Update $\bm \theta$ using gradient descent to minimize $\ell_{p}(\bm \theta)$
  \ENDFOR
\ENDFOR

\vspace{2pt}
\STATE \textit{$\triangleright$} \textbf{Stage I-B: Training $\bm f_{\bm \phi}$ on $\mathcal{D}_{\text{AI}}$}
\FOR{each epoch}
  \FOR{each mini-batch of images $\bm I \sim \mathcal{D}_{\text{AI}}$}
    \STATE Random crop $\tilde{\bm I}$ and compute $\bm x,\bm y,\bm x',\bm y'$ as in Stage I-A
    \STATE $\{\bm w_k,\bm \mu_k,\bm s_k\}_{k=1}^{K} \gets \bm f_{\bm \phi}(\bm x')$
    \STATE $q_{\bm \phi}(\bm y'|\bm x')=\prod_{(i,j)}\sum_{k=1}^{K} \bm w_k(i,j)\cdot \textsc{Logistic}\!\left(\bm y'(i,j)\mid \bm \mu_k(i,j), \bm s_k(i,j)\right)$
    \STATE $\ell_{\mathrm{NLL}}(\bm \phi)\gets -\log q_{\bm \phi}(\bm y'|\bm x')$ 
    \STATE Update $\bm \phi$ to minimize $\ell_{q}(\bm \phi)$
  \ENDFOR
\ENDFOR
\STATE Freeze $\bm \theta,\bm \phi$

\vspace{4pt}
\STATE \textit{$\triangleright$} \textbf{Stage II: Training classifier}
\STATE Initialize $\bm \psi$
\FOR{each epoch}
  \FOR{each mini-batch of labeled images $(\bm I,c)$, $\bm I \sim \mathcal{D}_{\text{photo}}\cup\mathcal{D}_{\text{AI}}$}
    \STATE Random crop $\tilde{\bm I}$ of size $s\times s$ from $\bm I$
    \STATE $\bm x \gets \bm M_R\odot \tilde{\bm I}_R + \bm M_G\odot \tilde{\bm I}_G + \bm M_B\odot \tilde{\bm I}_B$;\;\; $\bm x'\gets \textsc{Truncate}(\textsc{Stack}(\{\bm h_m * x\}_{m=1}^{M}),-t,t)$
    \STATE $\hat{c}\gets \bm g_{\bm \psi}(\textsc{Concat}(\bm f_{\bm \theta}(\bm x'),\bm f_{\bm \phi}(\bm x')))$
    \STATE $\ell_{\mathrm{cls}}(\bm \psi)\gets \textsc{BCE}(\hat{c},c)$ 
    \STATE Update $\bm \psi$ to minimize $\ell_{\mathrm{cls}}(\bm \psi)$
  \ENDFOR
\ENDFOR

\STATE \textbf{return} $\bm f_{\bm \theta},\bm f_{\bm \phi}, \bm g_{\bm \psi}$
\end{algorithmic}
\end{algorithm}

\begin{algorithm}[!ht]
\small
\caption{DCCT: Inference Phase}
\label{alg:dcct_infer}

\renewcommand{\algorithmicrequire}{\textbf{Input:}}
\renewcommand{\algorithmicensure}{\textbf{Output:}}
\begin{algorithmic}[1]
\REQUIRE
Test image $\bm I_{\mathrm{test}}$; trained $\bm f_{\bm \theta},\bm f_{\bm \phi}, \bm g_{\bm \psi}$;
CFA masks $\{\bm M_R,\bm M_G,\bm M_B\}$; high-pass filters $\mathcal{\bm H}=\{\bm h_m\}_{m=1}^{M}$; truncation threshold $t$;
Patch size $s$; number of test patches $P$; decision threshold $\tau$.
\ENSURE
Final score $\bar{s}$ and predicted label $\hat{c}$.

\FOR{$p=1$ to $P$}
  \STATE Random crop $\tilde{\bm I}_{\mathrm{test}}^{(p)}\in\mathbb{R}^{s\times s\times 3}$ from $\bm I_{\mathrm{test}}$
  \STATE $\bm x^{(p)} \gets \bm M_R\odot \tilde{\bm I}_R^{(p)} + \bm M_G\odot \tilde{\bm I}_G^{(p)} + M_B\odot \tilde{\bm I}_B^{(p)}$
  \STATE $\bm x'^{(p)} \gets \textsc{Truncate}(\textsc{Stack}(\{\bm h_m * \bm x^{(p)}\}_{m=1}^{M}),-t,t)$
  \STATE $s^{(p)} \gets \bm g_{\bm \psi}(\textsc{Concat}(\bm f_{\bm \theta}(x'^{(p)})),\bm f_{\bm \phi}(\bm x'^{(p)})))$
\ENDFOR
\STATE $\bar{s}\gets \frac{1}{P}\sum_{p=1}^{P} s^{(p)}$;\;\; $\hat{c}\gets \mathbbm{1}[\bar{s}>\tau]$
\STATE \textbf{return} $\bar{s},\hat{c}$
\end{algorithmic}
\end{algorithm}

\section{Further Evaluation of the One-Class Detector DCCT$^\dagger$}
In Sec.~\ref{subsec:further_analysis}, we introduced DCCT$^\dagger$, a one-class variant of our framework that is trained exclusively on photographic images and treats AI-generated content as an out-of-distribution anomaly. Concretely, DCCT$^\dagger$ leverages the same demosaicing-guided conditional model as DCCT, but replaces the binary classifier with a scalar anomaly score derived from the negative log-likelihood $\mathrm{NLL}{i,j}=-\log p_{\bm \theta}(\bm y_{i,j} \vert \bm x_{i,j})$ and the predictive entropy $H_{i,j}=-\sum_k p_{\bm \theta}(k \vert \bm x_{i,j})\log p_{\bm \theta}(k \vert \bm x_{i,j})$~\cite{cozzolino2024zero} of the learned color-correlation distribution,
and average these pixel-wise values to obtain an anomaly score: $D=\frac{1}{\mathcal{\vert I\vert}\mathcal{\vert J\vert}}\sum_{i=1}^{\mathcal{\vert I\vert}}\sum_{j=1}^{\mathcal{\vert J\vert}}(\mathrm{NLL}_{i,j}-H_{i,j})$,
where $\mathcal{\vert I\vert}$ and $\mathcal{\vert J\vert}$ are spatial height and width of an image, respectively.
In this appendix, we provide additional quantitative results, shown in Table~\ref{tab_genimage_dcct_oc}. We show that, despite never observing AI-generated images during training, DCCT$^\dagger$ achieves competitive detection performance, and even approaches or surpasses fully supervised binary detectors, highlighting the intrinsic separability induced by our ISP-aware pretraining task.

\section{High-Pass Filters}
\begin{figure*}[]
    \centering
    \includegraphics[page=15,trim=45mm 150mm 60mm 0mm,clip, width=0.9 \textwidth]{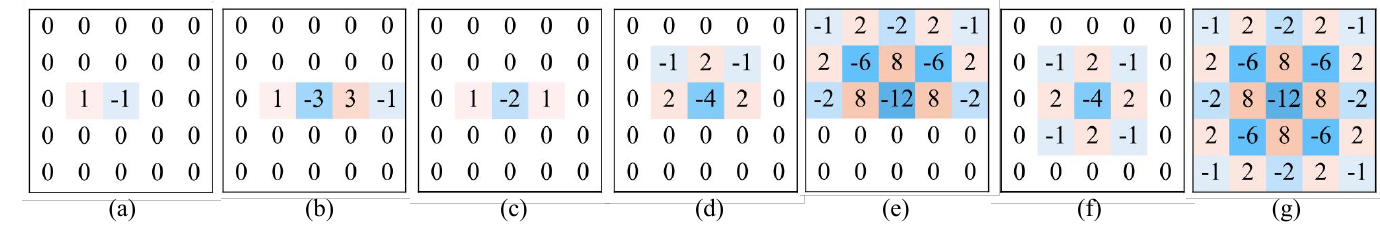}
    \caption{Seven prototype kernels for constructing the high-pass filter bank via discrete rotations.  (a) and (b) are rotated to eight compass directions $\{\nearrow, \rightarrow, \searrow, \downarrow, \swarrow, \leftarrow, \nwarrow, \uparrow\}$; (c) is rotated to four directions $\{\rightarrow, \downarrow, \nearrow, \searrow\}$ (opposite directions are equivalent); (d) and (e) are rotated to the four cardinal directions $\{\rightarrow, \downarrow, \leftarrow, \uparrow\}$; and (f) and (g) are used without rotation. In total, this yields $30$ high-pass filters ($2\times8 + 1\times4 + 2\times4 + 2$).}
    \label{fig_hpf}
    \vspace{-.3cm}
\end{figure*}

This section documents the high-pass filtering stage in DCCT. Following prior work in image forensics, we adopt the 30 high-pass filters originally proposed in SRM~\cite{fridrich2012rich}. All filters are implemented as 2D convolution kernels of size $5\times5$, making them straightforward to integrate into modern deep-learning frameworks (\eg, PyTorch) as standard convolutional layers. The full set of kernel coefficients is provided in Fig.~\ref{fig_hpf}. These filters are applied to the CFA-aligned single-channel inputs and corresponding target channels, yielding a 30-dimensional high-frequency residual representation per spatial location, which is then fed into the subsequent conditional modeling modules in DCCT.

\section{Example Images from Emerging Generative Models}
\begin{figure*}
    \centering
    \subfloat[GigaGAN~\cite{kang2023scaling}]{\includegraphics[width=0.9\textwidth]{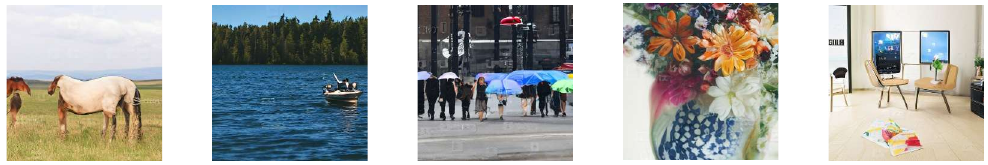}}\\
    \subfloat[DFGAN~\cite{tao2022df}]{\includegraphics[width=0.9\textwidth]{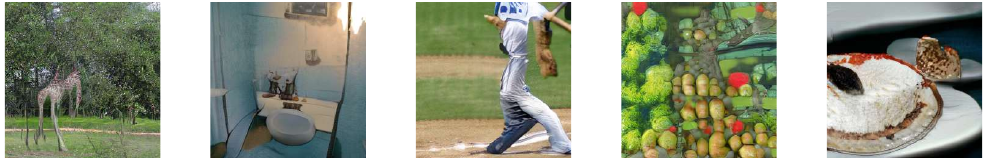}}\\
    \subfloat[GALIP~\cite{tao2023galip}]{\includegraphics[width=0.9\textwidth]{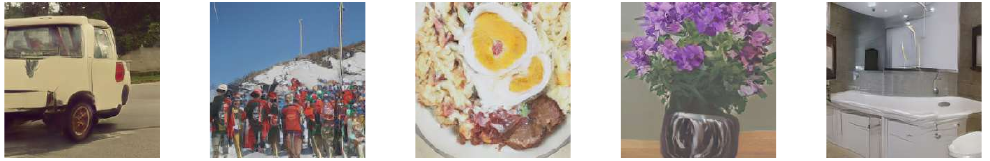}}\\
    \subfloat[FLUX.1~\cite{flux}]{\includegraphics[width=0.9\textwidth]{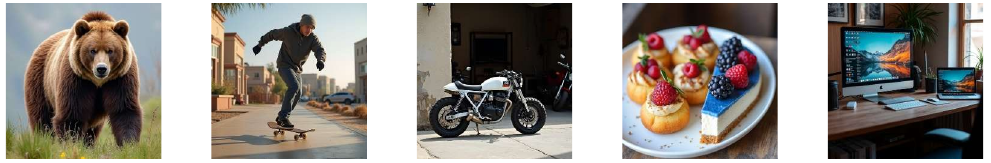}}\\
    \subfloat[FLUX.1-Kontext~\cite{flux}]{\includegraphics[width=0.9\textwidth]{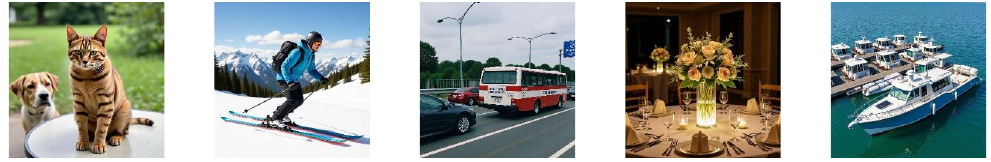}}\\
    \subfloat[SD-3.5-Turbo~\cite{sd3_5}]{\includegraphics[width=0.9\textwidth]{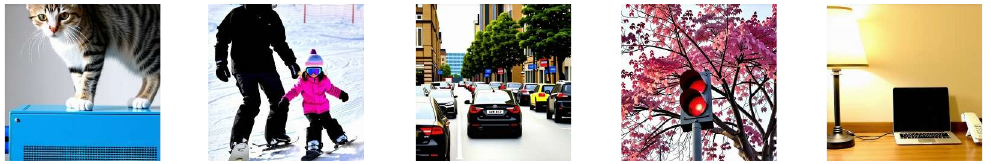}}\\
    \subfloat[Qwen-Image~\cite{wu2025qwen}]{\includegraphics[width=0.9\textwidth]{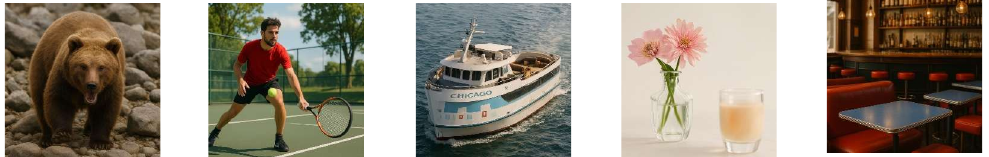}}\\
    
    \caption{Representative images from emerging generative models.}
    \label{fig:images_emerging_generators}
\end{figure*}

To illustrate the diversity and difficulty of the evaluation setting, this section presents representative examples of AI-generated images from several emerging generators that were not seen during training. As shown in Fig.~\ref{fig:images_emerging_generators}, these samples qualitatively demonstrate the visual similarity between photographic and AI-generated images, highlighting the challenges of reliable detection and the need to learn discriminative, generalizable features.

\end{document}